\pdfoutput=1

\documentclass[11pt]{article}

\usepackage[review]{emnlp2021}

\usepackage{times}
\usepackage{latexsym}

\usepackage[T1]{fontenc}

\usepackage[utf8]{inputenc}

\usepackage{microtype}

\usepackage{graphicx}
\usepackage{subcaption}
\usepackage{mwe}
\usepackage{xcolor}
\usepackage{booktabs}

\usepackage{enumitem}

\title{Frequency-based Distortions in Contextualized Word Embeddings}

\author{Kaitlyn Zhou \\ Stanford University \\ \texttt{katezhou@stanford.edu} 
\And
Kawin Ethayarajh\\ Stanford University \\ \texttt{kawin@stanford.edu} 
\And
Dan Jurafsky \\Stanford University \\ \texttt{jurafsky@stanford.edu}}

\begin{document}
\maketitle
\begin{abstract}
How does word frequency in pre-training data affect the behavior of similarity metrics in contextualized BERT embeddings? Are there systematic ways in which some word relationships are exaggerated or understated? In this work, we explore the geometric characteristics of contextualized word embeddings with two novel tools: (1) an \emph{identity probe} that predicts the identity of a word using its embedding; (2) the minimal bounding sphere for a word’s contextualized representations. Our results reveal that words of high and low frequency differ significantly with respect to their representational geometry. Such differences introduce \emph{distortions}: when compared to human judgments, point estimates of embedding similarity (e.g., cosine similarity) can over- or under-estimate the semantic similarity of two words, depending on the frequency of those words in the training data. This has downstream societal implications: BERT-Base has more trouble differentiating between South American and African countries than North American and European ones. We find that these distortions persist when using BERT-Multilingual, suggesting that they cannot be easily fixed with additional data, which in turn introduces new distortions.
\end{abstract}
\section{Introduction}

How similar are the words “felony” and “misdemeanor”? To a lawyer, the terms represent different charges, with different consequences, and different plans of action. To a layperson, these terms could be quite similar -- both referring to types of crime. Our perception of the similarity of words is conditioned on our backgrounds, and what is similar to one individual can be very different to another. One important variable is the frequency at which we are exposed to these concepts. In this paper, we study how the frequency of words in a model's training data plays a role in the semantic similarity of those words' contextualized embeddings. 

The impact of frequency has long been studied on static word embeddings \cite{levy2014neural, Hellrich_Hahn_2016, Mimno_Thompson_2017, Wendlandt_Kummerfeld_Mihalcea_2018}. For example, word frequency is known to affect the similarity between static embeddings, and is partially responsible for the existence of word analogies and bias in word2vec \cite{ethayarajh-etal-2019-towards, ethayarajh2019understanding}. But we know little about the effect of frequency on contextualized embeddings, where a word is represented not by a single point, but by a cloud of points. We'll call such a cloud \emph{sibling embeddings} or a \emph{sibling cohort}. Many questions arise from this change in representation. How can we characterize the space occupied by a word's sibling embeddings? What effect does a word's frequency in the training data have on its representational geometry? What downstream effects -- and potential social harms -- result from these differences?

To make sense of the space occupied by a word's sibling embeddings, we introduce two new tools: (1) an \emph{identity probe} that predicts the identity of the word using one of its sibling embeddings; (2) the minimal bounding hypersphere of a sibling cohort. The identity probe helps us measure how identifiable (i.e., linearly separable) a word is, and how this property is affected by its frequency, polysemy, and tokenization. Meanwhile, the radius of the bounding hypersphere gives us a measure of how diverse a word's representations are. Focusing on BERT \citep{devlin2018bert}, we find that the identifiability of a word -- as measured by the identity probe -- decreases with respect to its frequency. Conversely, the diversity of its representations -- as measured by the bounding hypersphere -- increases with respect to its frequency. This finding has both technical and social ramifications for how we use contextualized word embeddings.

Do these frequency-based geometric differences affect how we measure semantic similarity? Using the Stanford Contextualized Word Similarity dataset \cite{huang2012improving}, we find that the frequency of two words -- as manifested by the bounding hypersphere of both sibling cohorts -- helps explain the estimation error between cosine similarity and the human baseline (up to Pearson's $r = 0.42$). We call this effect a \textit{distortion} -- an over- or under-estimation -- in word similarity. 

How pervasive are these frequency-based distortions and do they affect topics with known biases and prejudices? We find that the names of countries from North America and Europe require larger minimal bounding spheres -- i.e., have more diverse representations -- than their counterparts from South America or Africa, correlating significantly with the country's gross domestic product. Furthermore, we find that North American and European country names are seen as more distinct from each other than country names from South America or Africa. 

In brief, we make three main findings:
\begin{itemize}
\itemsep0em
\item Word frequency in pre-training data affects the representational geometry of contextualized embeddings, with low frequency words being more identifiable and less diverse.
\item This leads to widespread distortions in word relations in the embedding space, especially with respect to semantic similarity.
\item These distortions disproportionately affect under-represented populations, such as those in South America and Africa.
\end{itemize}

Finally, we argue that although distortions can be harmful, there is no such thing as a neutral or undistorted model; rather each corpus has its own distortions, and awareness of these differences is critical.

\section{Methods and Data}
We used the March 1, 2020 Wikimedia Download and BookCorpus frequencies \cite{zhu2015aligning, hartmann-dos-santos-2018-nilc} to approximate word frequencies in the BERT pre-training corpus. We refer to this as a \textit{word's training data frequency}. We use sentences from English Wikipedia to extract word embeddings by averaging the last four hidden layers of BERT. Short (less than 20 characters) and long (over 512 characters) sentences are filtered out. We explored alternative methods of creating word embeddings, such as various ways of concatenating layers, but they produced almost identical results. We matched keywords on token IDs to ensure punctuation and casing are consistent across examples. In the case where words were split into multiple parts from the tokenizers, the embedding of the first token was used. Implications of this decision will be discussed in depth in section \ref{section: tokenization paragraph}.

\section{BERT Embedding Identifiability}\label{ssec:identifiability}

Static embeddings occupy a single, distinct point in space, but the contextualized embeddings for a word occupy a cloud instead. How do we characterize the geometric representation of contextualized word embeddings? How do these characteristics vary with a word's training data frequency? 

\subsection{Identity Probe}
First we introduce the \textbf{identity probe}, a simple logistic regression classifier that is trained with few-shot learning to take an instance of a sibling embedding, and predict which word it comes from. Since BERT was trained to predict the identity of masked tokens in context, the identity probe is testing the model on a task that is related to its training. By studying for which words this classification decision is easy, and for which it is hard, we can study how lexical properties influence the latent properties of words and concepts in embedding space.

\paragraph{Experimental Details} 
Our experiments look at the performance of the identity probe on a random sample of 24,000 words. We measure how this performance varies across word characteristics like word frequency (\# of occurrences in BERT's original training data), word sense (from Word Net), and word tokenization (based on BERT's tokenizer).\footnote{We started with a list of 30,000 words then filtered out words that had non A-Z or a-z characters and words that were rare (less than 100 occurrences). We wanted all models to have 1000 classes and rounded the sample size down to 24,000 and took a random sample of that size from our remaining words.}

Rather than training a 24,000-class classifier, we simplify the task and built twenty-four 1000-class classifiers. Each class is a word, and class selection for each model was random. This design decision allows the models to overall perform with high accuracy (~90\%) despite their limited complexity, suggesting that the results reflect discrepancies in word embeddings rather than poor model performance.\footnote{Larger models (such as 5,000-class classifiers) were built. However, they performed worse and required significantly more computational time and examples to train.} We use one-shot learning at training time and test models on 10,000 examples (ten examples per class). Since most words perform very well on the identity probe, we focus on the words where the model made errors in its classification.\footnote{We used the logistic regression model from the scikit-learn library using a one-vs-rest (OvR) scheme.} 

\begin{figure*}[t]
     \centering
     \begin{subfigure}[b]{\columnwidth}
         \centering
         \includegraphics[width=\textwidth,height=4.8cm]{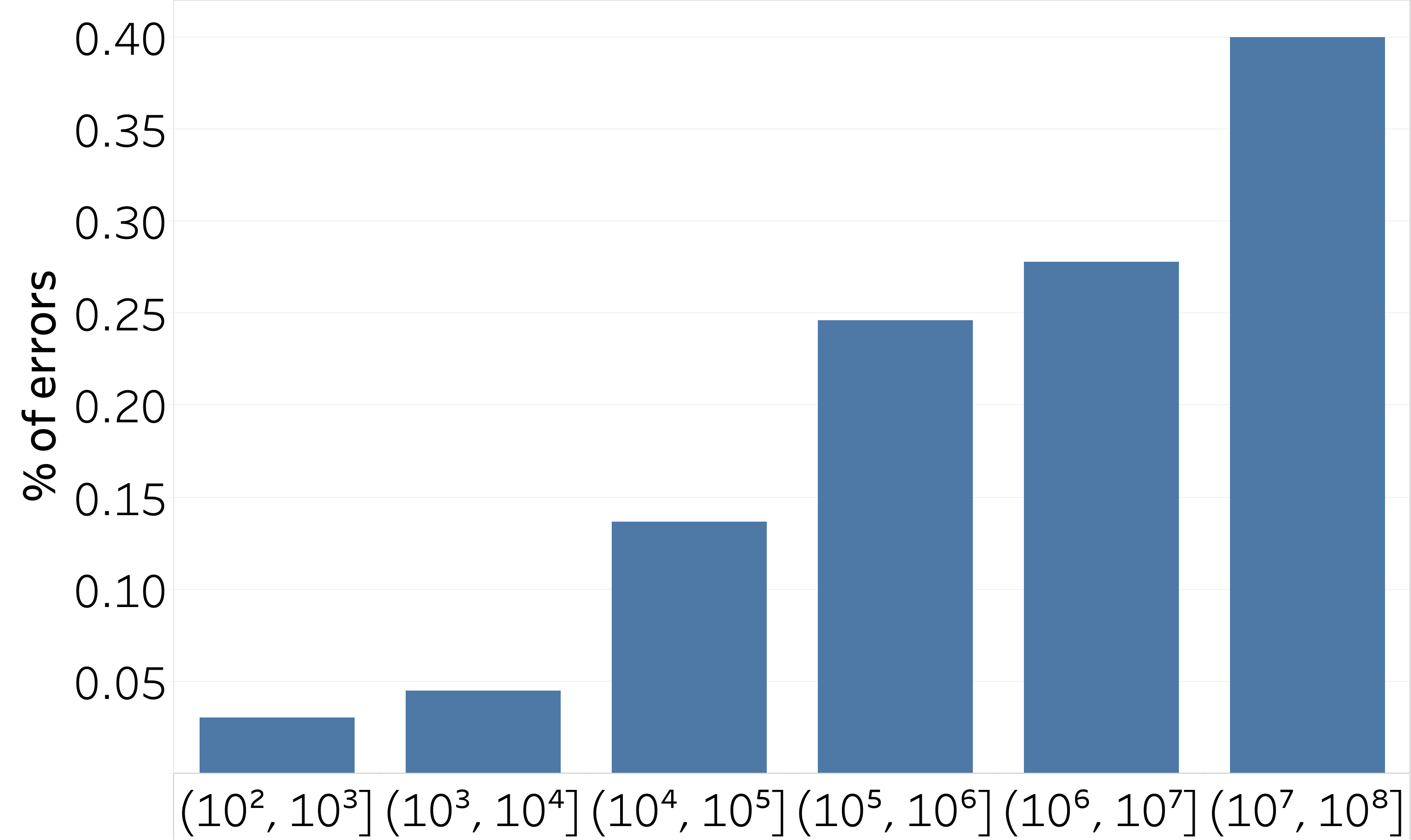}
         \caption{}
         \label{fig:word frequencies}
     \end{subfigure}
     \begin{subfigure}[b]{\columnwidth}
         \centering
         \includegraphics[width=\textwidth,height=4.8cm]{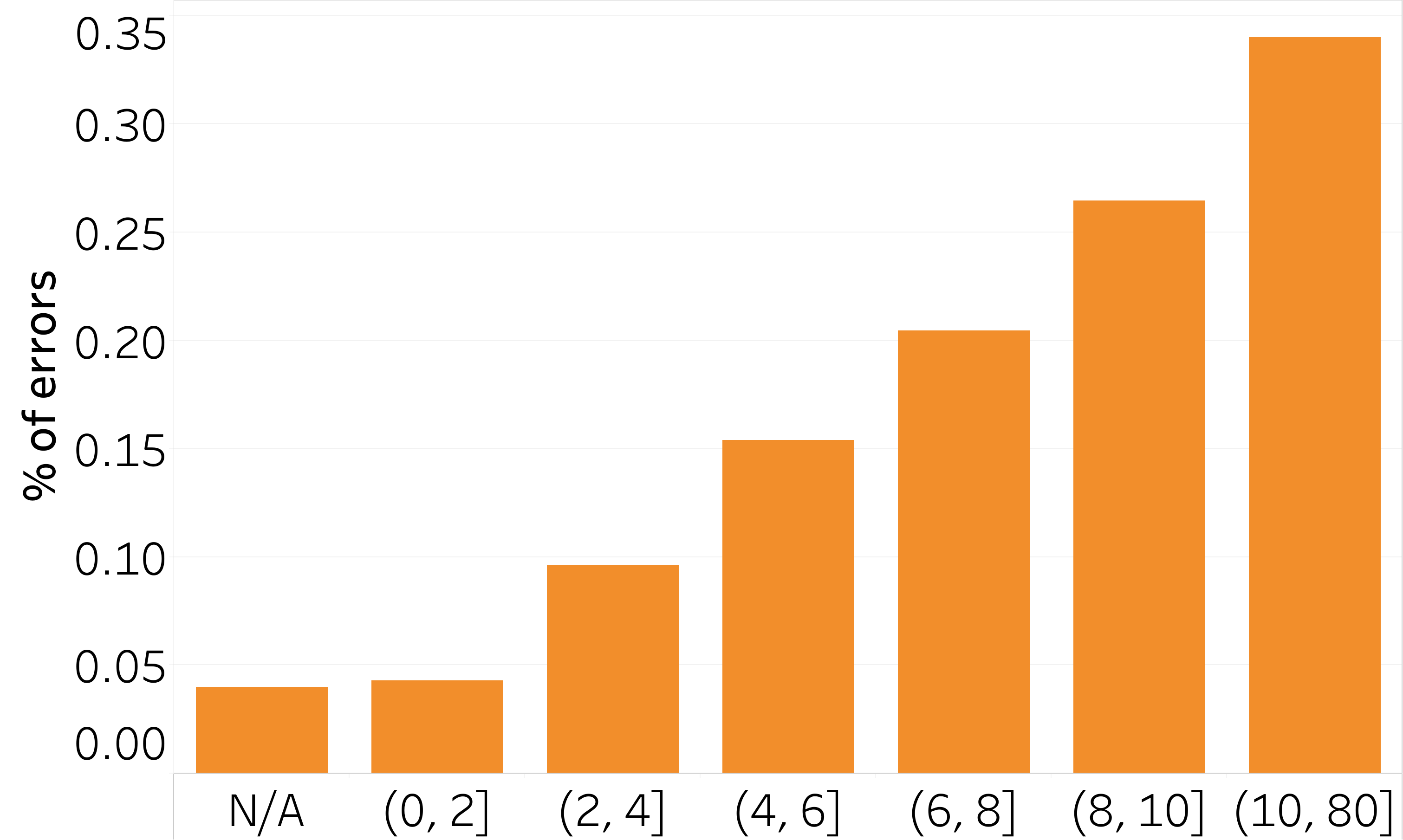}
         \caption{}
         \label{fig: word senses}
     \end{subfigure}
     \begin{subfigure}[b]{\columnwidth}
         \centering
         \includegraphics[width=\textwidth,height=4.8cm]{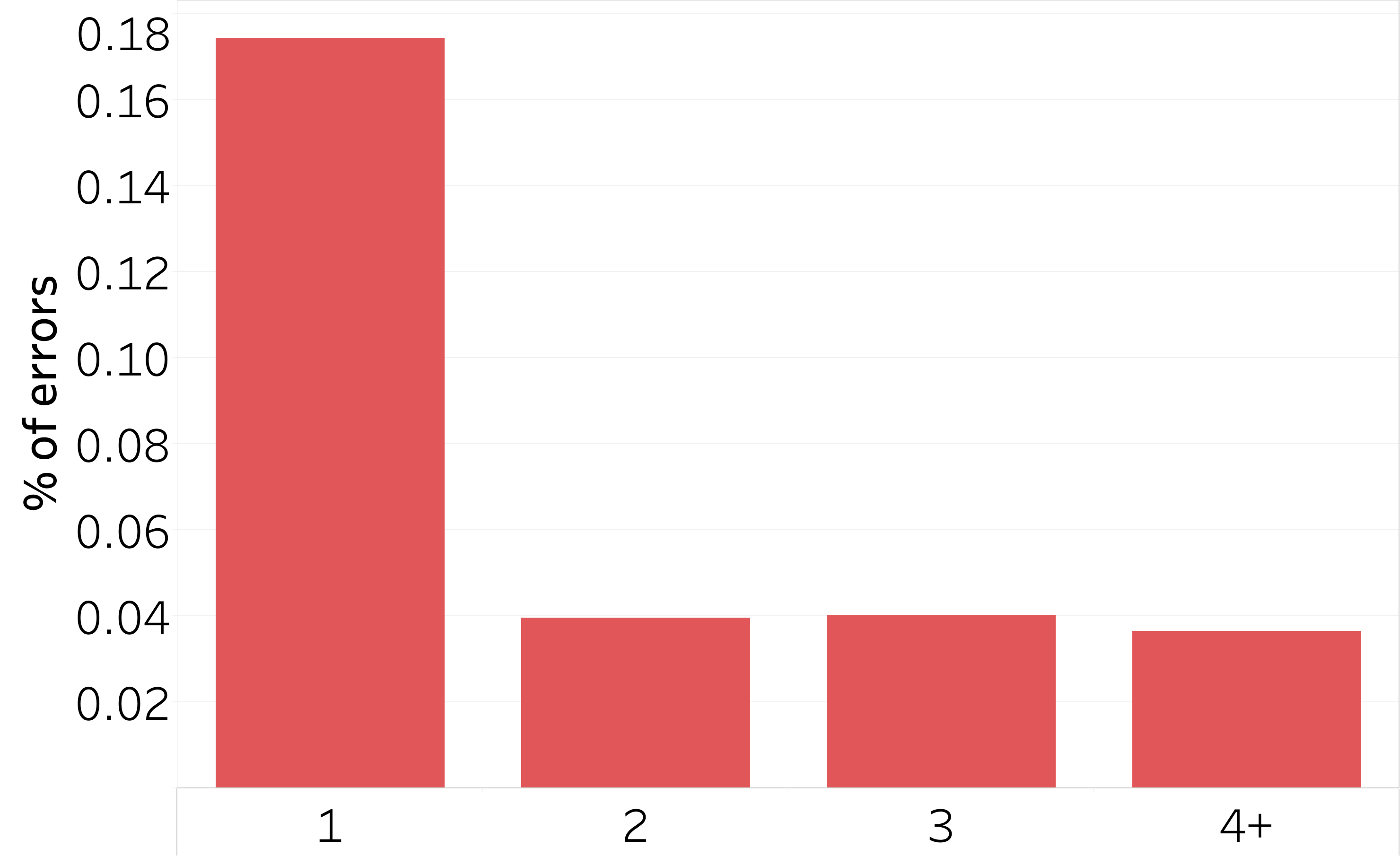}
         \caption{}
         \label{fig: number of tokens}
     \end{subfigure}
     \begin{subfigure}[b]{\columnwidth}
         \centering
         \includegraphics[width=\textwidth,height=4.8cm]{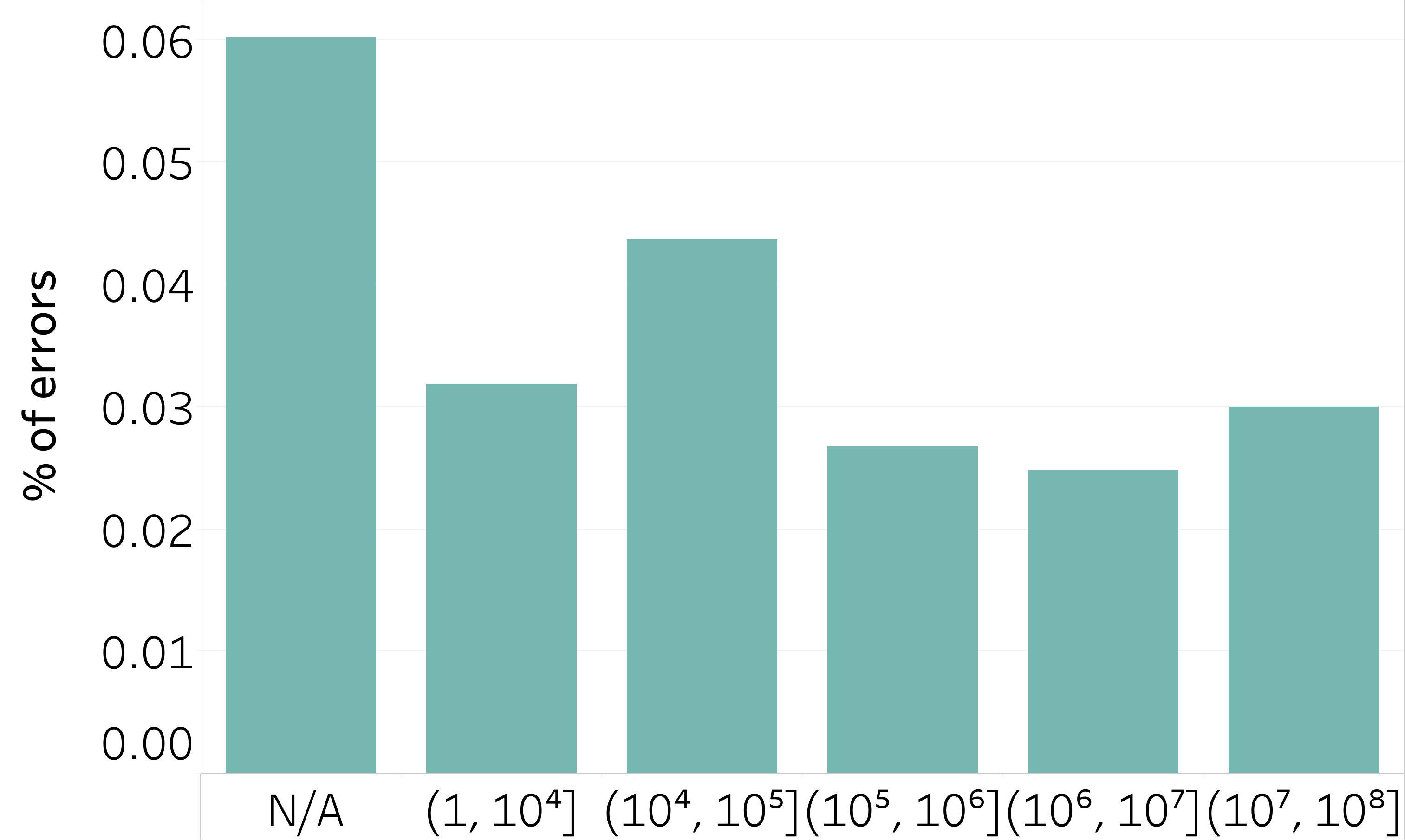}
         \caption{}
         \label{fig: frequency of first token}
     \end{subfigure}
     \caption{The bar charts above highlight the percentage of errors for words binned by frequency, senses, or tokens. (a) \% of errors in words binned by word frequencies (b) \% of errors in words binned by \# of word senses (c) \% of errors in words binned by number of tokens, (d) \% of errors in words binned by frequency of the first token}
     \label{fig:Identity Probe Results}
     \vspace{-0.5 em}
\end{figure*}

\paragraph{Frequency} We find that a word's training data frequency correlates negatively with identifiability i.e., linear separability in the embedding space, given our linear probe. The accuracy of the identity probe decreases as word frequency increases. For example, the error rate of words with over 10 million training data occurrences is 40\%, compared to an error rate of 3\% for rare words, which occurred between 100 and 1000 times in the training data (see figure \ref{fig:word frequencies}). Although this may seem surprising, it concurs with past work finding that stopwords -- a subset of highly frequent words -- have highly context-specific BERT embeddings \citep{Ethayarajh_2019}.

One explanation for the poor performance of high-frequency words could be the high polysemy of these words \cite{zipf1945meaning}. Indeed, the identity probe makes more errors with polysemous words. Very polysemous words (more than 10 senses in WordNet) are 8 times more likely than monosemous words to be misidentified (34\% versus 4\%, see figure \ref{fig: word senses}). Another explanation for lower linear separability of high frequency words is that the space occupied by high frequency words tends to be larger  -- i.e., the embeddings are more spread out. This would most likely lead to difficulty in identifying them with a simple logistic regression model. This hypothesis is explored in section \ref{section:minimum bounding hyperspheres}, on minimum bounding hyperspheres. 

\paragraph{Tokenization} 
\label{section: tokenization paragraph}
Does our decision to represent a word via the embedding of its first token impact a word’s identifiability? We find that this is largely not the case. BERT-Base has a 30,000 token vocabulary, with words that occurred over approximately 10,000 times in its original training data considered in the vocabulary. Thus the word “intermission”, which is out-of-vocabulary, is tokenized into “inter” and “\#\#mission”, and we would use the (extremely ambiguous) first token “inter” to represent “intermission”. 

Surprisingly, using only the first token to represent an OOV word had little impact on the identifiability of words, suggesting that these embeddings could capture enough context to differentiate themselves from words with identical prefixes. We find that words tokenized into multiple pieces had lower error rates (4\%) than words that remained whole (17\%) (see figure \ref{fig: number of tokens}). In other words, the words “intermission”, “interpromotional”, “interwar”, and “interwoven” are distinguishable from one another even though each are tokenized into “inter” and subsequent tokens and only the first token’s embedding is used. That is, the context (namely, the subsequent token “\#\#mission”) sufficiently changed the BERT embedding for “inter” to make it identifiable in context. The fact that words which were not split into multiple tokens performed worse as a group is likely explained by our prior finding that high frequency words have lower performance in the identity probe, since words that were single tokens were in vocabulary and have a higher frequency.

We find that the performance of the identity probe degrades slightly when the word is tokenized into multiple parts and the first token maps to a non-word (a single or two characters, excluding words like “a”, and “i”). For example, the identity probe does slightly worse on words such as “bleach”, which is tokenized into “b”, “\#\#lea”, “\#\#ch”. The error rate of these words is 6\% compared to 2 - 4\% error rate across all other categories (see figure \ref{fig: frequency of first token}).

\subsection{Minimum Bounding Hyperspheres}
\label{section:minimum bounding hyperspheres}

\begin{figure}[ht]
    \centering
    \vspace{-0.5 em}
     \includegraphics[width=.48\textwidth]{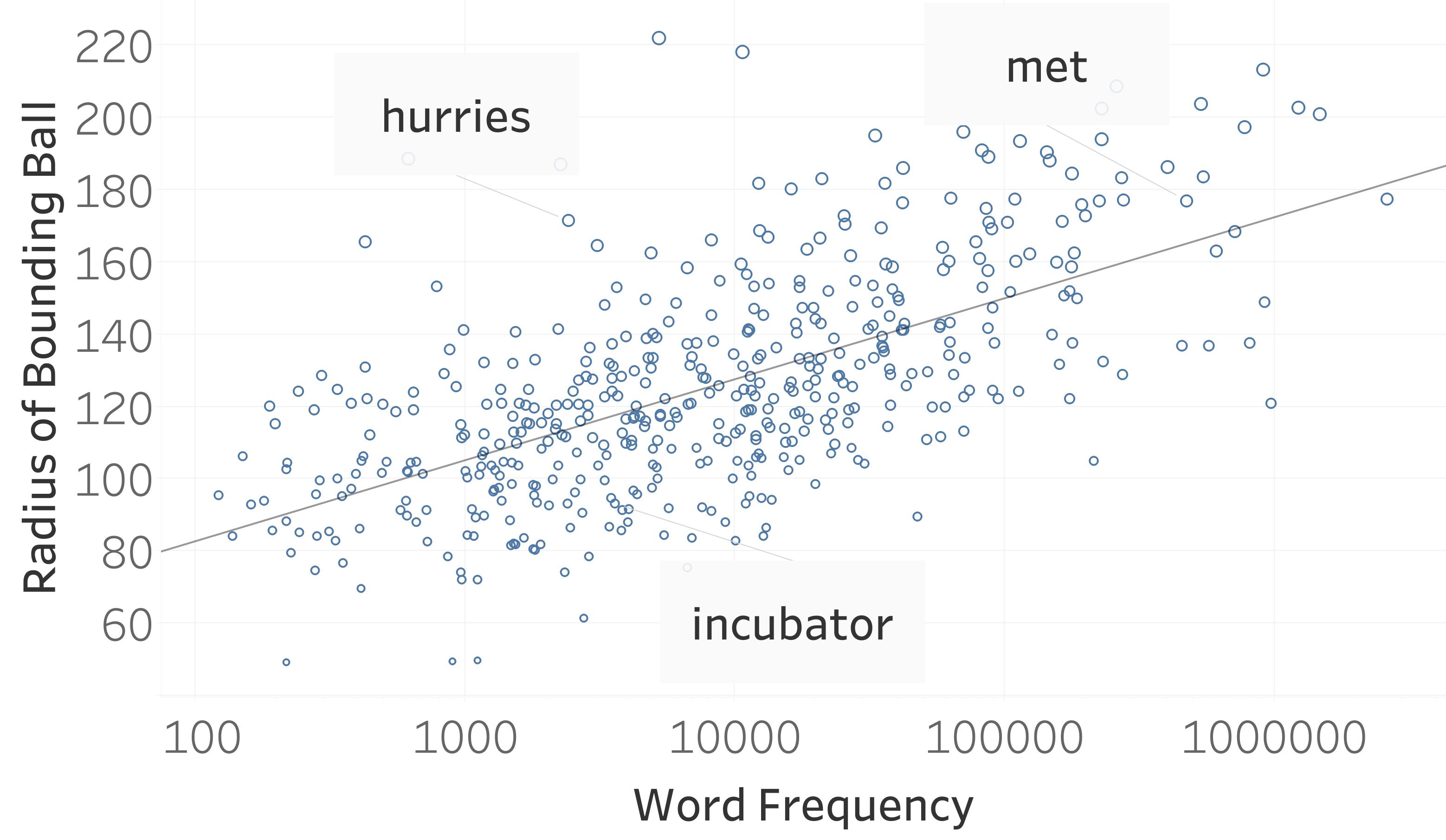}
     \vspace{-1 em}
     \caption{The radius of the minimal bounding ball of sibling embeddings of words is correlated with log(word frequency). Pearson's r: 0.63, \textit{p}  < .001}
     \label{fig: radius vs word frequency}
     \vspace{-0.5 em}
\end{figure}

How can we characterize the space occupied by a cloud of contextualized word embeddings? We use the minimum bounding hypersphere (a.k.a., bounding ball) of a word cohort, which encompasses the space occupied by sibling embeddings as the second method to study the embedding geometry. There are many ways to measure the space created by high-dimensional vectors; here we use the radius of the $n$-ball that encompasses all the sibling embeddings. Our results are robust to various other measures of wordspace, including the average pairwise distance between sibling embeddings and taking the PCA of these vectors and calculating the convex hull of sibling embeddings in lower dimensions. 

For a sample of 500 words, for each word we took 10 instances of its sibling embeddings and calculated the average radius of its bounding balls. The radius of bounding balls was normally distributed with a Shapiro test statistic of 0.99 (\textit{p} < .001) with a standard deviation of 1.27. We find that unlike static embeddings -- where word embeddings occupy the same amount of space (a single point) -- the space occupied by contextualized embeddings varies greatly depending on the word's training data frequency. The more frequent the word, the larger the radius of the bounding ball (Pearson's r: 0.63, \textit{p} < .001) (see figure \ref{fig: radius vs word frequency}). Since the radius was calculated in 768 dimensions, it’s important to note that an increase in just 1\% of the radius results in a volume of the bounding sphere that is nearly 2084 times bigger in size.

\paragraph{} Our first experiments illustrate that certain words have embeddings that can easily identify them, whereas other words are much more difficult to identify by their embeddings. Part of these results are explained by words' training data frequencies. Additionally, we find that the space occupied by a sibling cohort varies with the frequency of the word in its original training data.

\section{Semantic Distortions}

\begin{figure*}[t]
    \centering
    \begin{subfigure}[b]{\columnwidth}
        \centering
        \includegraphics[width=\textwidth]{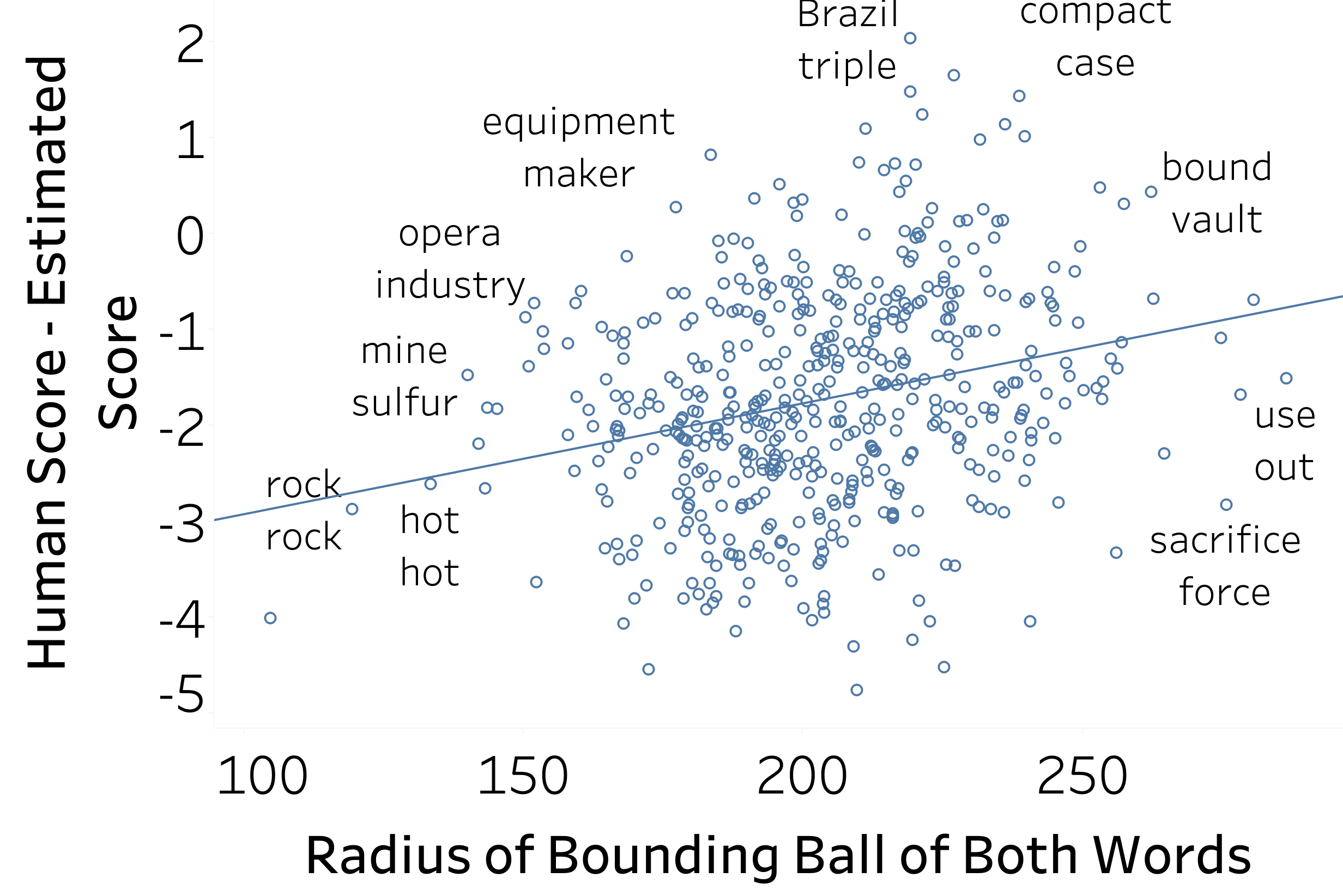}
        \caption{}
    \end{subfigure}
    \begin{subfigure}[b]{\columnwidth}  
        \centering 
        \includegraphics[width=\textwidth]{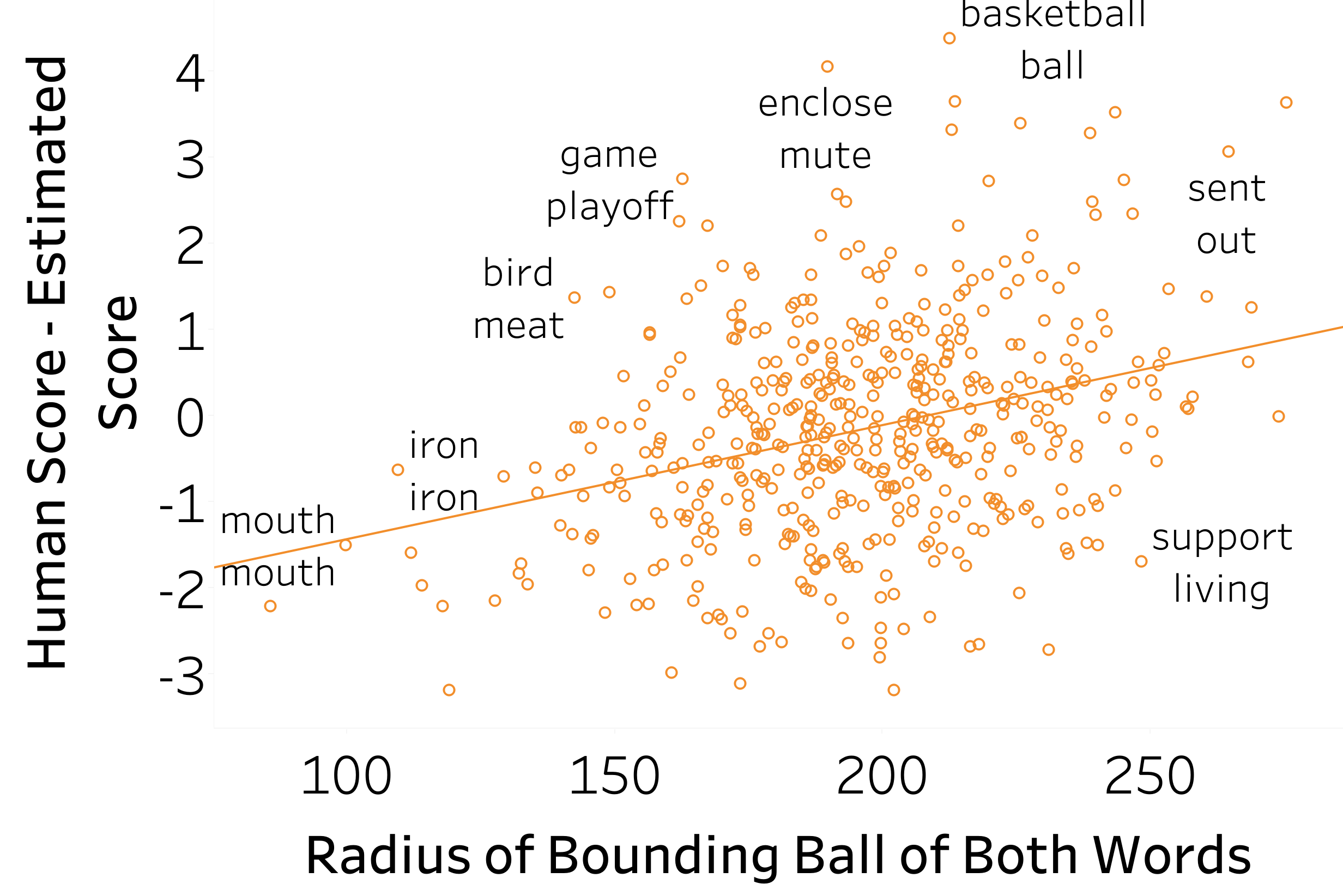}
        \caption{}    
    \end{subfigure}
    \begin{subfigure}[b]{\columnwidth}   
        \centering 
        \includegraphics[width=\textwidth]{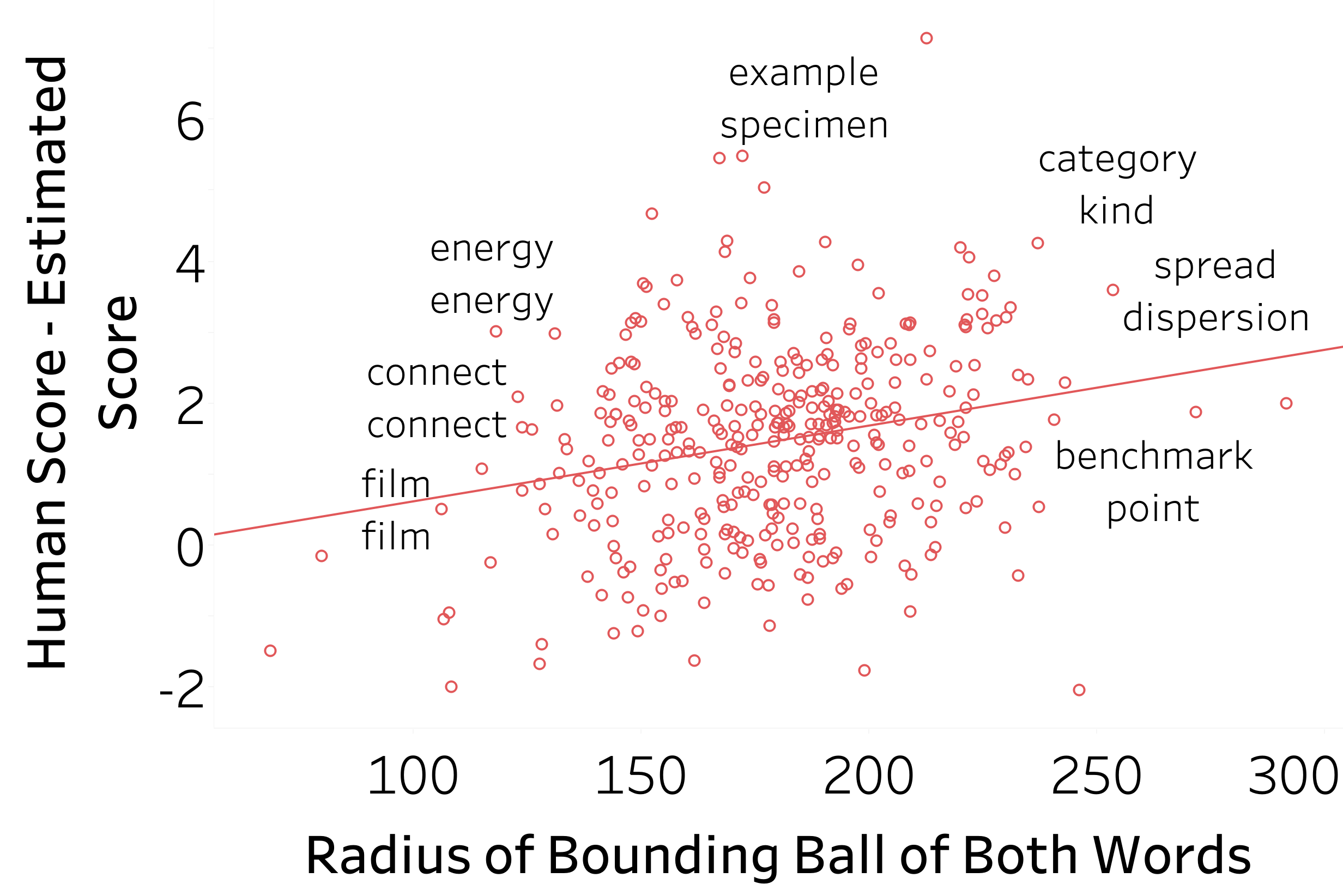}
        \caption{}   
    \end{subfigure}
    \begin{subfigure}[b]{\columnwidth}   
        \centering 
        \includegraphics[width=\textwidth]{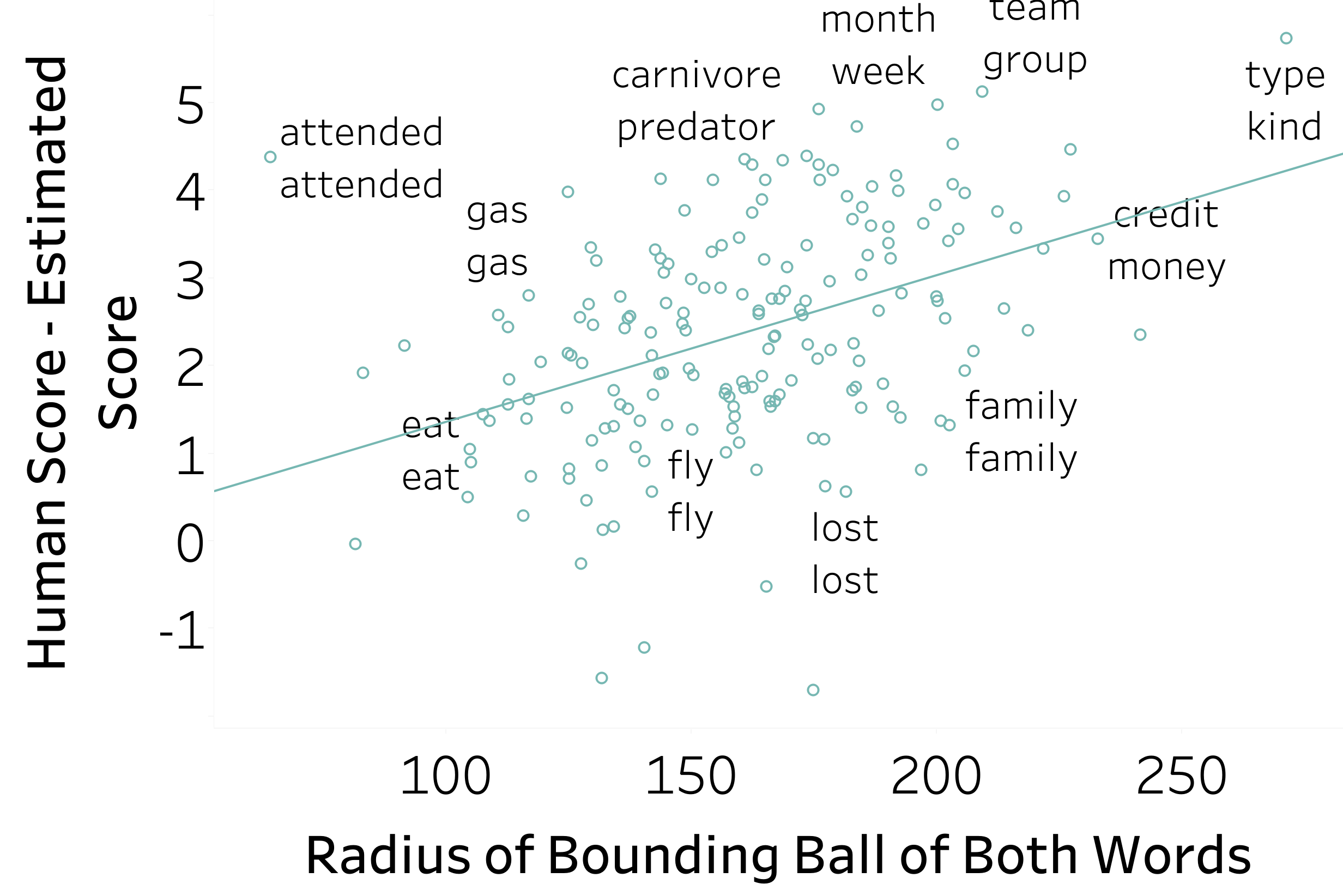}
        \caption{}    
    \end{subfigure}
    \caption{A positive correlation between the size of the word spaces and the difference in human and cosine similarity. Pearson's r = (a) 0.26 (first quartile, pairs that are the least similar to each other), (b) 0.31 (second quartile), (c) 0.24 (third quartile), (d) 0.42 (fourth quartile, pairs that are the most similar to each other); \textit{p }< .001. The larger the radii of the bounding balls, the bigger the difference between the human and predicted scores. We charted by quartile to adjust for heteroscedasticity. This difference also correlates with word frequency but has a weaker relationship, Pearson's r between 0.10 and 0.20, \textit{p} < 0.02 } 
    \label{fig:scws results}
    \vspace{-1 em}
\end{figure*}

Is there a harm from certain sibling cohorts having larger minimum bounding balls than others? What, if any, are the effects on semantic similarity of having frequency-based geometric differences in embeddings? We turn to a standard dataset with human annotations on the semantic similarity of words in context and we measure if semantic similarity, as calculated using cosine distance, is distorted in relation to the word's training data frequency. 

\subsection{Empirical Comparison to Human Similarity}

We test whether the \textbf{difference} between cosine similarity and human-judged similarity of two words can be explained by the radius of the bounding ball that encompasses both and find that this is indeed the case. We then present theoretical intuitions for why, when using cosine distance, larger bounding balls yield a wider distribution of distances.

We use the Stanford Contextualized Word Similarity Dataset \cite{huang2012improving} as a human baseline for the similarity between two words. The dataset was collected through averaging ten human annotations of pairs of words in sentential context, with 0 being the least similar and 10 being the most similar. For each pair of words used in the dataset, we then calculated the cosine similarity between embeddings for the two words in the sentential context they were evaluated in. 

Studying the relationship between human-judged and BERT-based similarity required that we account for two facts about the data. Firstly, cosine similarity and human annotation rankings have a Spearman correlation of 0.59. To account for this, we use linear regression to predict human annotations using cosine similarity scores. This predicted similarity is then compared to the human scores as a way to account for the correlation between human annotations and cosine similarity. Secondly, the difference between cosine similarity and human annotations is heteroscedastic: as human-judged similarity increases, the \textbf{difference} between the predicted and actual score does as well. The second factor was mitigated by splitting the dataset into quartiles based on the human annotations, with the fourth quartile being the words that are most similar to each other.

The results show a correlation between the difference in predicted and actual similarity and the size of these word spaces. The bigger the radius of the bounding balls of the two words, the more likely the cosine similarity is to be smaller than the human-judged similarity. The smaller the space, the more likely cosine similarity is to be bigger than the human-judged similarity. For example the words “alcohol” and “chemistry” have one of the smallest radii (127.7) and have a higher predicted similarity (5.8) than the human annotated similarity (3.6). On the other hand, “eliminate” and “out” have one of the largest radii (275.6) but have a much smaller predicted similarity (0.76) than human annotated similarity (4.4). In this case, cosine similarity yields very different results for two pairs of words that according to humans, are comparably similar. This error correlates (up to Pearson's r: 0.42) with the size of the bounding balls that encompass these embeddings; words whose bounding balls are larger are seen as far more distinct than those whose balls are smaller (see figure \ref{fig:scws results}).

\subsection{Theoretical Intuition}
Here we offer some theoretical intuition for why using cosine distance to estimate semantic similarity can lead to over- or under-estimation. First, consider the 2D case. 
\begin{itemize}
    \itemsep0em 
    \item Let the center of the bounding ball be $x_c$. 
    \item Let $w$ denote the normalized target word vector, against which we're measuring cosine distance. 
\end{itemize}

If we normalize every point in the bounding ball, it will form an arc on the unit circle. The length of this arc is $2\theta = 2 \arcsin \frac{r}{\|x_c\|_2}$:

\begin{itemize}
    \itemsep0em 
    \item Let $\theta$ denote the angle made by $x_c$ and a vector tangent to the bounding ball that passes through the origin.
    \item $\sin \theta = \frac{r}{\|x_c\|_2}$, so the arc length on the unit circle is $r \theta = \arcsin \frac{r}{\|x_c\|_2}$ (normalized points) 
    \item  Multiply by 2 to get the arclength between both (normalized) tangent vectors.
    \item Let $\ell_w$ denote the arclength between $w$ and the closest edge of the arc, without loss of generality.
\end{itemize}

If $w$ is not on the arc, then the arclength between $w$ and a point on the arc lies in $[\ell_w, \ell_w + 2\theta]$. If $w$ is on the arc, then it lies in $[0, 2\theta - \ell_w]$. 

In both cases, the range of arclengths is monotonic increasing in $\theta$. Since $\theta$ is monontonic increasing in $r$, we get by transitivity that the range of arclengths is monotonically increasing in the radius of the bounding ball. Note that the cosine of the arclength between two unit vectors is their cosine similarity (aka dot product because they're normalized). Since the possible arclengths between $w$ and points on the arc must fall within $[0, \pi]$, and cosine is monotonic decreasing on this interval, the range of cosine similarities between $w$ and points on the arc is monotonic decreasing in $r$. Since the cosine distance is $1 - \cos(x,y)$, the range of cosine distances is monotonic increasing in $r$. In other words, holding constant the center of the bounding ball, bigger bounding balls lead to greater range in the cosine distance. Although this does not necessarily hold if you change $x_c$ while also changing the radius, because contextualized word representations are highly anisotropic \citep{Ethayarajh_2019}, the impact of any such change in $x_c$ may be limited.

Together with our empirical results, this finding has large-scale implications for downstream tasks, given that single-point similarity metrics are used in a variety of methods and experiments \cite{reimers2019sentence, reif2019visualizing}. The unequal frequency of words in the pre-training data can lead to over- or under-estimation of semantic similarity. 

\section{Geographic Distortions}
How pervasive are distortions of semantic similarity? Do these distortions cause societal harm, particularly on topics that have biases and prejudices? We drew on the well-known journalistic bias in which journalists treat the African continent as a single homogeneous entity, as if African countries are less differentiated than countries in wealthier regions \cite{Nothias_2018}. We hypothesize that BERT may show similar effects; because some country names are more prevalent in English Wikipedia and English Books, they may have larger bounding balls, leading to semantic similarity distortions across country names.

\begin{center}
\begin{table*}[ht]
\begin{tabular}{lrrrrrrrr}
\toprule
Region & \multicolumn{1}{r}{\begin{tabular}[r]{@{}r@{}}North\\ America\end{tabular}} & \multicolumn{1}{r}{Europe} & \multicolumn{1}{r}{\begin{tabular}[r]{@{}r@{}}Middle\\ East\end{tabular}} & \multicolumn{1}{r}{Asia} &  \multicolumn{1}{r}{\begin{tabular}[r]{@{}r@{}}South\\ America\end{tabular}} & \multicolumn{1}{r}{Oceania} & \multicolumn{1}{r}{\begin{tabular}[r]{@{}r@{}}Central\\ America\end{tabular}} & \multicolumn{1}{r}{Africa} \\\midrule
BERT-Base & 100\% & 92\%  & 92\% & 91\% & 89\%  & 87\% & 85\% & 85\% \\ \midrule
\begin{tabular}[l]{@{}l@{}}BERT-Base\\ (Artificial Dataset)\end{tabular} & 100\%  & 89\% & 92\% & 89\% & 88\%  & 88\% & 87\% & 87\% \\ \midrule
BERT-Multilingual & 100\%& 89\%  & 88\% & 88\% & 91\%  & 81\%  & 83\%  & 83\%  \\ \bottomrule
\end{tabular}
\caption{\% of the average radius of bounding balls relative to the average of radius of bounding balls of North American countries names. Central America also includes countries in the Caribbean.}
\label{table: radius_countries}
\end{table*}
\end{center}
\subsection{Anglo- and Euro-centrism}
To start, we find that for cities with a population greater than 100,000 people, 39\% of North American cities and 27\% of European cities are in BERT-Base’s vocabulary but less than 6\% of city names from all other continents are in BERT’s vocabulary. Contrast Nigeria -- which has the world’s fourth largest population of English speakers (after the United States, India, and Pakistan) -- with the United Kingdom \footnote{The city and population data was created by MaxMind, available from http://www.maxmind.com/"}. There are 92 Nigerian cities with a population of 100K+, but only two ($\approx 2\%$) of them are in BERT’s vocabulary. In contrast, the United Kingdom has 46 large cities in the vocabulary, out of a possible 67 ($\approx 69\%$). The percentage of cities in and out of vocabulary provides context for the distribution of topics in BERT's pre-training data.

\begin{figure}[ht]
    \centering
    \includegraphics[width=.48\textwidth]{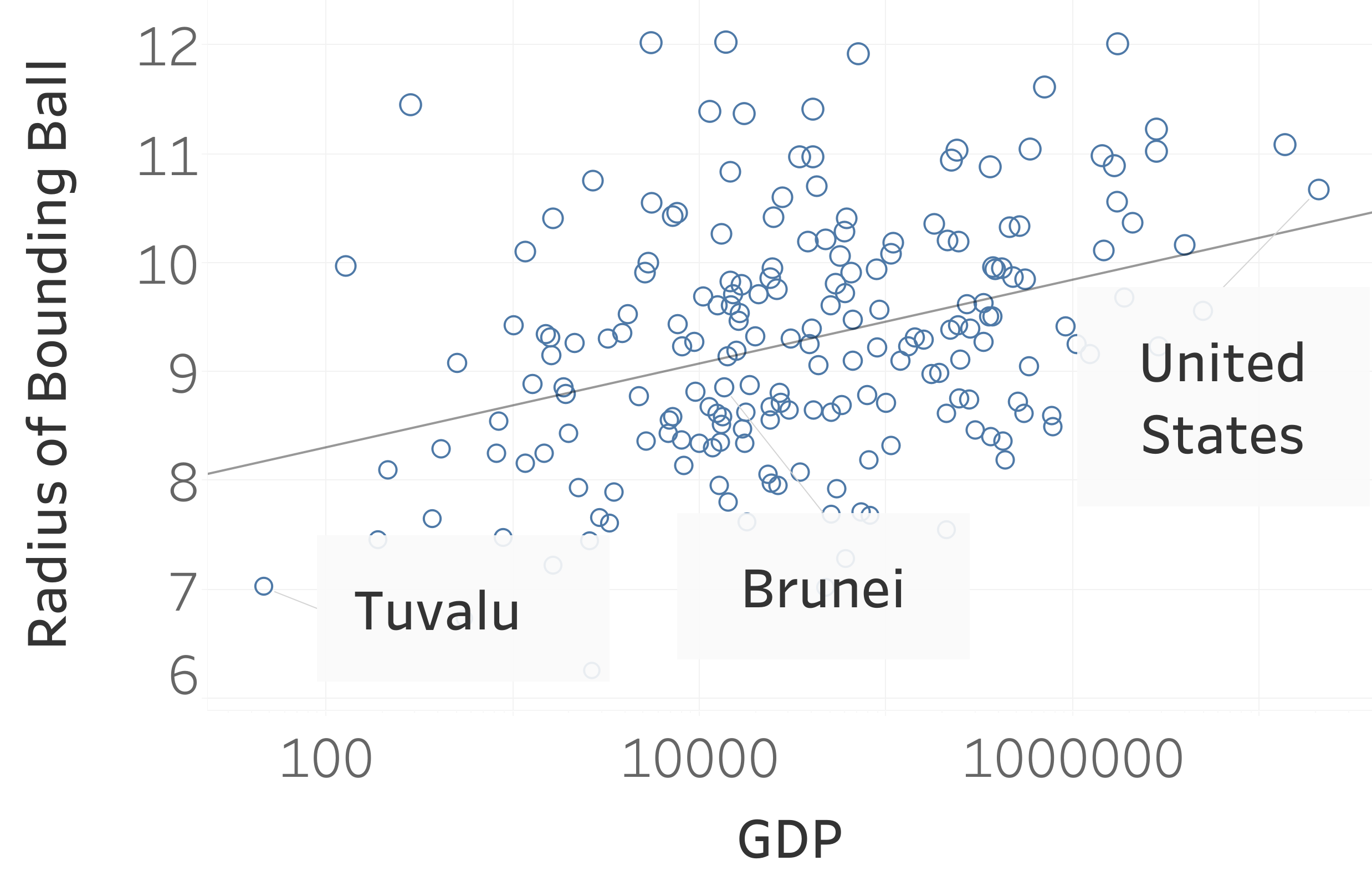}
    \vspace{-2 em}
    \caption{Radius of minimum bounding balls with respect to log(Gross Domestic Product) (Pearson's r: 0.36, \textit{p} < 0.001).}
    \label{fig:gdp by radius}
    \vspace{-0.5 em}
\end{figure}

This unequal distribution of geographic information results in larger bounding balls for certain countries. Notably, North American countries have the largest bounding balls and countries from Africa, Central America and the Caribbean, and Oceania have the smallest, by nearly 15\% (see table \ref{table: radius_countries}).\footnote{Names of countries were checked for polysemy and names with other frequent senses were manually inspected and filtered out such as: “Chad”, “Jordan” “Mali”, Georgia” and “Guinea” (differentiating from “New Guinea”). Other polysemous countries' names such as "China" and "Turkey" were not filtered as case matching would have eliminated references to “china” or “turkey”.} A pattern can also be seen when comparing the size of the radius of the minimal bounding ball with the Gross Domestic Product (GDP) of each country (Pearson's r: 0.36, \textit{p} < 0.001). Countries (n= 53) with GDPs of less than 10 billion dollars on average have bounding balls that are 19\% smaller than the bounding balls of countries with the largest GDPs (United States and China) (see figure \ref{fig:gdp by radius}).  

\begin{table}[ht]
\begin{tabular}{llr}
\toprule
\multicolumn{1}{l}{Region} & Country & \begin{tabular}[c]{@{}r@{}}\# similar \\ countries\end{tabular} \\\midrule
North America & United States & 1 \\ \midrule
North America & Canada & 6 \\ \midrule
North America & Mexico & 11 \\ \midrule
Europe & France & 12 \\\midrule
Europe & Germany & 19 \\\midrule
Europe & Spain & 25 \\\midrule
South America & Peru & 23 \\\midrule
South America & Guyana & 29 \\\midrule
South America & Venezuela & 51 \\\midrule
Africa & Uganda & 30 \\\midrule
Africa & Kenya & 31 \\\midrule
Africa & Algeria & 37 \\
\bottomrule
\end{tabular}
\caption{Examples of average number of countries that shared a cosine similarity of 0.7 or greater with the country in question. For example, across ten contextualized instances of "Canada" compared to ten contextualized instances of all other country names, 6 country names were found to have a cosine similarity of 0.7 or greater with "Canada".}
\label{table:cosine similarity country}
\vspace{-0.5 em}
\end{table}

One possible confounding result in these experiments is whether these differences in bounding balls and semantic similarity are due to context, some country names appearing in more diverse contexts than others. To account for this, we designed an artificial dataset, named BERT-Base Artificial, where country names were substituted into identical contexts (e.g., "This year I want to visit [country name] and try their local cuisine."). The results illustrate that even in identical contexts, certain regions still had larger bounding ball radii and are seen as more distinct than others. African, Oceanian, and Central American and the Caribbean countries on average have 10-15\% smaller radii than North American countries (shown in table \ref{table: radius_countries}). 

How do these geometric differences in bounding balls cause distortions in practice? Semantic distortions arise when we use standard metrics such as cosine distance as a way to calculate similarity between country names. Specifically, North American countries are seen as more distinct from all other 197 countries as compared to South American or African Countries. Table \ref{table:cosine similarity country} gives an example of this finding. When the United States is compared to all other country names, it shares a cosine similarity of 0.7 or higher with only 1 country, whereas Venezuela shares a cosine similarity of 0.7 or higher with 59 other countries. These results represent stark differences in the semantic similarity seen across country names, with certain countries seen as more distinct than others. These results show the same trend when using word embeddings from the artificial dataset as well, which would control for any confounding factors in the diversity of context of country names. On average, North American country names have 5 countries with a cosine similarity of 0.7 or higher but South American country names have on average 31 countries that have a cosine similarity of 0.7 or higher. For additional rigor and concerns of how multi-word country names would affect these results, the same experiments were run with only single-word country names with nearly identical results.  

\subsection{Controlling for Context but Distortions Persist}
How could these distortions persist when words are in identical contexts? One hypothesis for why we see distortions even in artificial datasets is that BERT has learned -- from its training data -- to embed more information about the context into the representation of high frequency words. This richer contextual information encoded in high-frequency words would explain why the difference persists even when the context of the keywords is identical. Our appendix includes an additional experiment that tests for this exact phenomenon with a larger sample of random words. We find similar results: when placed in identical contexts, words of higher frequency capture more information about their context in their word embeddings than lower frequency words.

\subsection{Additional Data Fails to Mitigate the Effects of Distortions}
If the distortions we have studied are caused by word frequency in the training data, could the effects of these distortions be mitigated by adding more training data? We tested this hypothesis by looking at the word distortions and semantic similarities when using BERT-Multilingual (BERT-ML), which is trained with the same original data as BERT-base with the addition of Wikipedia articles from 104 other languages.

The relationships between names of countries do change in the BERT-Multilingual embedding space, highlighting that it is possible to change these distortions with changes in pre-training data. Unfortunately, the embeddings do not improve many of the inequities that exist (see table \ref{table: radius_countries}). This could possibly be explained by the fact that although BERT-Multilingual includes many more languages, the amount of data added was not proportional. We see the same patterns as with regular BERT, both in the size of the bounding balls and with respect to which countries are seen as most distinct (i.e., North American and European countries remain at the top). This calls into question whether these distortions can be mitigated in this way; since all datasets (even balanced ones) will still have frequency differences and distortions.

\section{Discussion and Conclusion}
The diversity of contextualized representations differs across words, as shown through differences in the identifiability and minimal bounding spheres of their embeddings. More frequent words are less identifiable and less clustered in embedding space. As a result, when using canonical metrics of similarity such as cosine distance, words with high frequency are more likely to be seen as less similar as compared to the human baselines, highlighting a distortion in word relationships. This imbalance of word and topic frequency ultimately reflects an Anglocentric and Eurocentric view of the world. Although these word relations can change with different training data -- at least in the domain of geography -- the training data added to BERT-Multilingual do little to mitigate the effects of the distortions we identified.

If the ground truth is the semantic relationship between entities in the real world, all datasets will have some sort of distortion, as they are not grounded in the real world, but only in its textual depiction \cite{glenberg00,lake2020word,bisk20,bender2020climbing}. But even if there were a perfectly neutral dataset  -- where topics occurred in equal frequency and biases around how each topic was written about were neutral or non-existent -- it would still have a distortion, just one where some relationships are the same. 

There is no set of distortions that is ideal or perfect. Additional transparency around training data, such as word frequency, could help bring awareness to the potential inequalities that exist in datasets and, by extension, the models that are trained on them. As previous literature has suggested \cite{gebru2018datasheets, mitchell2019model,bender2018data}, documentation for datasets is critical for increasing transparency and accountability in machine learning models. There is a lack of awareness that frequency-based distortions exist and we recommend that word frequencies and distortions in models be revealed to users, similar to the model limitations and biases listed by organizations like HuggingFace.\footnote{https://huggingface.co/bert-base-cased} Lastly, research efforts that focus on how to fine-tune models to take on a variety of distortions could give rise to new perspectives and help provide more diverse and grounded representations of the world.

\section*{Acknowledgements}
We thank Dallas Card, Nelson Liu, Amelia Hardy, and Tianyi Zhang for their helpful feedback and discussion! Kaitlyn Zhou is supported by a Junglee Corporation Stanford Graduate Fellowship. This work was supported by a SAIL-Toyota Research Award. Toyota Research Institute (“TRI”) provided funds to assist the authors with their research but this article solely reflects the opinions and conclusions of its authors and not TRI or any other Toyota entity.
\bibliography{custom}
\bibliographystyle{acl_natbib}

\appendix

\section{Appendix}
\label{sec:appendix}

For additional validation for the results from section 6, we ask: are there differences in how much context is embedded into words and does this vary with frequency? We design an experiment called, \textbf{contextual retrieval}, where the task is to identify a word's original context when given only the word's embedding. The words are embedded all in the same contexts by using an artificial dataset of sentences (n = 30) where we randomly select an index to insert the target word. For training, we replace the random word with BERT's mask token ("[MASK]") and create embeddings of the masked token in all the different contexts. The labels correspond to the sentence the word embedding was created with. At test time, we give the model embeddings of word in one of the contexts. When testing with a set of five hundred different random words, a simple logistic regression model achieves an accuracy of 97\%.

\begin{figure}[h]
    \centering
    \vspace{-0.5 em}
     \includegraphics[width=.48\textwidth]{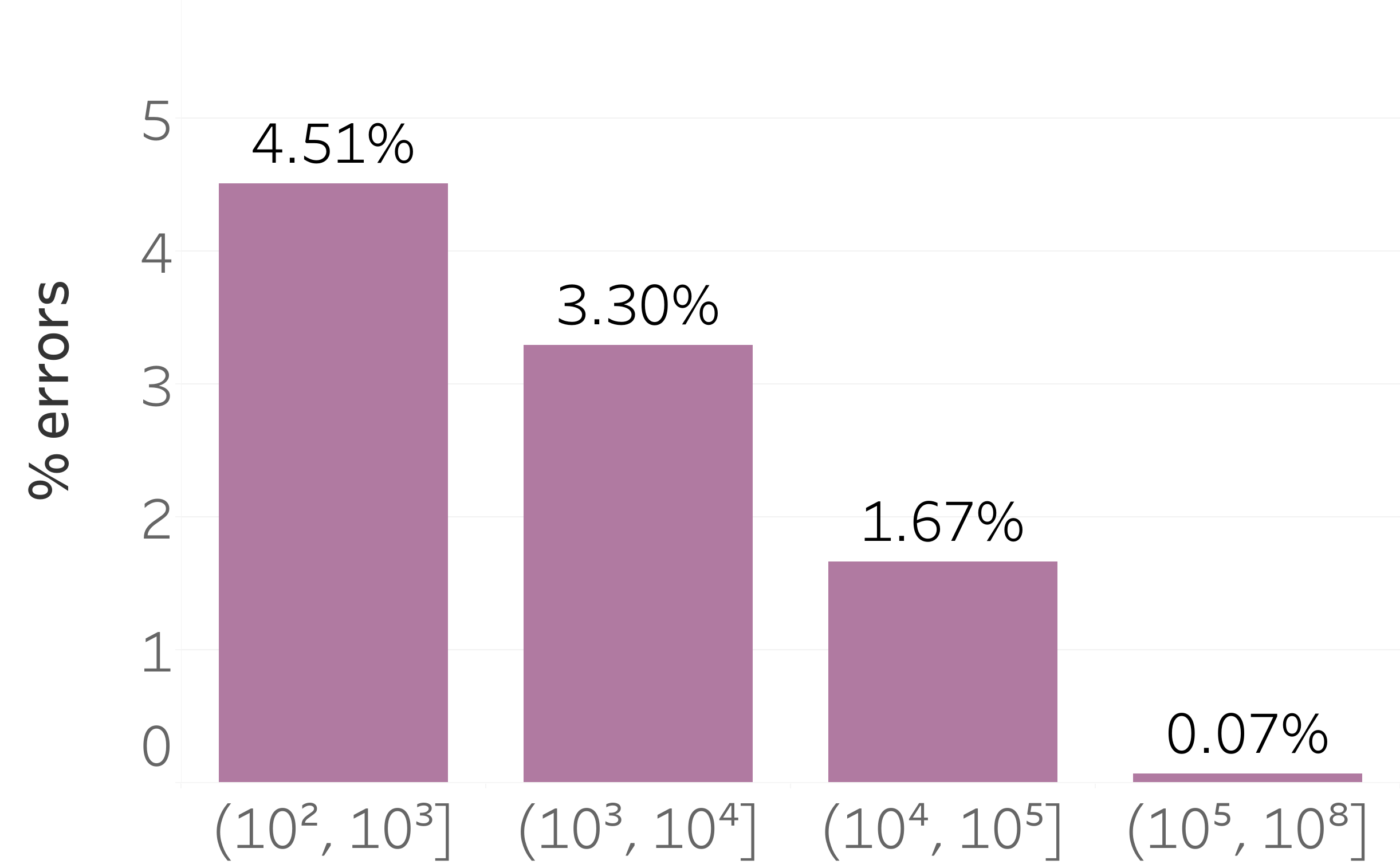}
     \vspace{-1 em}
     \caption{The error rate is negatively correlated with the frequency of the word in BERT's original training data.}
     \label{fig: contextual retrieval}
     \vspace{-0.5 em}
\end{figure}

We find that it is easy to retrieve the original context of a word given its word embedding and that there is a correlation (Pearson's r: 0.36, \textit{p} < 0.001) between the log(frequency) of the word in BERT's original training data and the performance of the word on this task. We see that words with lower frequency (< 1,000 occurrences in training data) have a 4.5\% error rate whereas higher frequency words (> 100,000 occurrences in training data), have less than a 0.07\% error rate (see figure \ref{fig: contextual retrieval}).

This experiment highlights the discrepancies in the amount of context that is embedded and how it varies with a word's training data frequency. It supports our hypothesis that frequency-based distortions are partly due to how the model embeds high and low frequency words differently.

\end{document}